# Unbiasing on the Fly: Explanation-Guided Human Oversight of Machine Learning System Decisions

*Mamman, Hussaini[1,2], Basri, Shuib[1], Balogun, Abdullateef O.[1], Imam, Abdullahi Abubakar[3], Kumar, Ganesh[1], and Capretz, Luiz Fernando[4,5]

[1] Department of Computer and Information Sciences, Universiti Teknologi PETRONAS, Bandar Seri Iskandar, 32610 Perak, Malaysia
`{*hussaini_21000736,shuib_basri,abdullateef.ob, ganesh_17005106}@utp.edu.my`

[2] Department of Management and Information Technology, Abubakar Tafawa Balewa University, Bauchi, 740272, Nigeria
`mhussaini@atbu.edu.ng`

[3] School of Digital Sciences, Universiti Brunei Darussalam, BE1410, Brunei Darussalam
`abdullahi.imam@ubd.edu.bn`

[4] Dept. of Electrical & Computer Eng., Western University, London, Canada
`lcapretz@uwo.ca`

[5] Visiting Professor, MCS Program, Yale-NUS College, 138527, Singapore

**Abstract.** The widespread adoption of ML systems across critical domains like hiring, finance, and healthcare raises growing concerns about their potential for discriminatory decision-making based on protected attributes. While efforts to ensure fairness during development are crucial, they leave deployed ML systems vulnerable to potentially exhibiting discrimination during their operations. To address this gap, we propose a novel framework for on-the-fly tracking and correction of discrimination in deployed ML systems. Leveraging counterfactual explanations, the framework continuously monitors the predictions made by an ML system and flags discriminatory outcomes. When flagged, post-hoc explanations related to the original prediction and the counterfactual alternatives are presented to a human reviewer for real-time intervention. This human-in-the-loop approach empowers reviewers to accept or override the ML system decision, enabling fair and responsible ML operation under dynamic settings. While further work is needed for validation and refinement, this framework offers a promising avenue for mitigating discrimination and building trust in ML systems deployed in a wide range of domains.

**Keywords:** Fairness testing, online testing, counterfactual explanations, human review

## 1 Introduction

Machine learning (ML) systems are now commonly used to make autonomous decisions in various critical areas of human lives. ML systems are used to decide who gets hired for a job [1, 2], whose loan is approved by a bank [3, 4] or which products are bought [5]. They are also used to decide which diagnoses and treatments a patient receives [6] and who gets bailed or sentenced [7, 8]. The use of ML systems can reduce human error, increase efficiency and scalability, and enhance the accuracy of decision-making processes [9]. However, with the increasing use of ML systems in human-focused areas, there is growing concern about the fairness of their decisions [10, 11]. For example, in 2018, Reuters uncovered age and gender bias in Amazon's resume screening algorithm, which downgraded resumes containing words like "women's" and graduation years suggesting older applicants. Amazon swiftly scrapped the biased algorithm, but likely deployed it for months without realizing its unfair impacts. To minimise such ethical concerns, it is imperative to continue to test the fairness of ML systems before and after their deployment to prevent causing harm to people during their operations.

Recently, various efforts have been made in the software engineering community to first search for and mitigate discrimination in ML, borrowing from conventional software testing techniques. For example, Udeshi et al. (Udeshi et al. 2018) introduced AEQUITAS, a fairness testing technique that explores the input space of the ML model for discriminatory instances and then perturbs the non-protected attributes of those instances to generate more discriminatory samples. Monjezi et al. [13] presented DICE which uses causal algorithms to identify discriminatory elements by measuring the use of protected information in the decision-making of neurons in deep neural networks (DNNs). DICE then suggests a mitigation strategy involving intervention in the implicated neurons. While current approaches have proven effective, they primarily concentrate on detecting and addressing discrimination in the development phase. There is comparatively less emphasis on testing for fairness during the operation of a deployed ML system [14–16]. To address this gap, this study proposes a framework for continuous monitoring of a deployed ML system to assess its fairness.

The proposed framework provides a way to continuously monitor the predictions of an ML system and alerts when discrimination is detected. The framework generates counterfactual examples of an input instance received by the system for making predictions. Counterfactual examples are alternative scenarios that are created by applying a minimum perturbation to the features of an instance that can cause its prediction to flip. When counterfactuals are found, they are presented, along with the original instance, to a human reviewer to make a decision. The post-hoc explanation is provided to further assist the human reviewer in understanding how the model makes its judgment on the original instance and the counterfactuals. Unlike traditional one-time fairness testing, our proposed framework adapts to changing data and usage patterns, proactively alerting human reviewers to potential biases before they harm users.

The rest of this paper is organised in the following manner: Section 2 gives a background of the concept of fairness and fairness testing. Section 3 reviews the current state of fairness testing research. Section 4 introduces our methodology, which details the proposed conceptual model. Section 5 showcases practical applications of our model with real-world examples. Finally, Section 6 offers concluding remarks and outlines plans for future work.

## 2  Background

This section will start with an overview of essential terminologies before delving into fairness, fairness testing, counterfactual explanations, and human review.

### 2.1  Fairness

Fairness in the ML system's decision is the absence of bias or preference toward a person or group based on their inherent or acquired attributes [17]. Formulating fairness is the first step in resolving fairness issues and developing a fair ML model. In simple terms, let $X$ represent a set of individuals and $Y$ represent the true label set when making decisions about each individual in $X$. The ML predictive model, represented by $h$, has been trained using a dataset that includes sensitive attributes, denoted by $S$, and other attributes, defined by $Z$,

$$F(X,Y,h,S,Z) = fair\ if\ h(X,S,Z) \approx h(X,S',Z), \qquad (1)$$

Where $S'$ represents a subset of $S$ that has been modified to remove any potential bias. The formulation can also be written as a probability distribution:

$$F(h) = Pr[Y = y | X, A = a] = Pr[Y = y | X, A = a'] \qquad (2)$$

for all $y \in Y, a, a' \in A, and\ X$ It defines fairness as the equality of the probability of a specific outcome ($y$) given an individual's attributes ($X$) and a specific value of the protected attribute ($a$) and the likelihood of the same result given the same individual's attributes and a different value of the protected attribute ($a'$). This definition states that the model is considered fair if the predicted outcome ($y$) is independent of the protected attributes ($a$) for the same individual ($X$).

### 2.2  Individual fairness

Individual fairness ensures that similar individuals receive similar outcomes regardless of their protected attributes, such as race or gender [18]. This type of fairness is based on the idea that individuals should be treated regardless of their membership in a particular group. Let $\mathbb{Q} = \{q_1, q_2, \cdots, q_n\}$ represent a collection of attributes (or features), where $n$ is a natural number and $n \in \mathbb{N}$. The $i$-$th$ attribute in $\mathbb{Q}$ is denoted as $q_i$. Each attribute $q_i (\in \mathbb{Q})$ is associated with a set of values, known as the domain of $q_i$, represented by $D(q_i)$. These domains $D(q_i)_{i \in n}$ are distinct from each other. The input space $X$ for a set of attributes $\mathbb{Q}$ is defined as the Cartesian product of the domains of $(q_1, q_2, \cdots, q_n \in \mathbb{Q})$, expressed as $X = D(q_1) \times D(q_2) \times \cdots \times D(q_n)$. A constituent $x\ of\ X$ is considered a data item, also known as an input instance. We denote the value of the $i$-$th$ attribute of input $x \in X$ as $x_i$. Additionally, we introduce $Q_{protected} \subseteq \mathbb{Q}$ as the set of protected attributes (e.g., race, age, gender, education level). In our interpretation, an ML classifier with input space $X$ is represented as a function $h$; hence, we use $h(x)$ to signify the outcome or decision made by the trained classifier $h$ for the input $x$.

### 2.3 Fairness Testing

Fairness testing is a branch of software testing that focuses on activities intended to expose fairness bugs in ML systems [15]. A fairness bug is any flaw in an ML system that causes a conflict between the obtained and required fairness conditions [19]. The existence of fairness bugs resulting in an unfair decision by ML systems motivates research into various fairness testing techniques that can help detect these bugs [15].

Fairness testing can be classified as offline and online testing. Offline fairness testing is done during the development of a model. It only evaluates the fairness of a model based on the training data it was given, without considering the model's performance in real-world situations where unseen inputs are inevitable [20]. In contrast, online fairness testing involves continuously monitoring and evaluating the fairness of a deployed ML system online, using real-world input data [20].

Online fairness testing of ML systems in real-world scenarios is of utmost importance for maintaining fairness in their decision-making processes. This evaluation provides valuable information on the system's real-time performance. It facilitates the detection of any biases that may have infiltrated its decision-making over time, enabling prompt corrective action to be taken to ensure fairness [15, 20].

### 2.4 Counterfactual Explanations

Counterfactual explanations for ML models are a technique for identifying the minor changes necessary to alter a given prediction. This is done by providing a specific data instance that is similar to the observation under examination but yields a different result. The goal is to understand the direct impact of certain factors on the outcome, which can help identify potential problems with the model and make informed decisions.

Given an input feature $x$ and its corresponding output from model $f$, a counterfactual explanation is a change made to the input that would lead to a different output $y$ from the same model. This approach allows users to see the direct impact of certain factors on the outcome, which can assist them in identifying potential problems with the model to make informed decisions. The following formulation explains the concept [21, 22].

$$c = \arg\min_{c} y\text{loss}\,(f(c), y) + |x - c| \qquad (3)$$

The first term, $yloss(f(c), y)$, represents the loss or error in the model's prediction for a specific data point c. The function $f(c)$ represents the model's prediction for the data point $c$, and $y$ represents that data point's true or correct label. The loss function $yloss(f(c), y)$ measures the difference between the model's prediction and the actual label. The second term, $|x - c|$, represents the difference between the original observation $x$ and the counterfactual data point $c$. This term ensures that the counterfactual data point c is like the initial observation $x$, but still results in a different prediction. In summary, the first part ($yloss$) aims to move the counterfactual example

$c$ towards a prediction that is different from the original input. In contrast, the second part ensures that the counterfactual example remains like the initial input.

Kusner et al. [23] proposed the concept of counterfactual fairness, which tackles bias in ML models by considering the causal relationships within the data. The causal generative model provides a more thorough examination of potential discrimination in the model, preventing instances of bias towards specific individuals or groups from going unnoticed and offering a more comprehensive understanding of the model's discriminatory behaviour [24]. Let $a$ represent a protected attribute, $á$ represent the counterfactual attribute of $a$, and $x'_i$ represent the new input with $a$ replaced by $a'$. Model $h$ is counterfactually fair if, for each input $x_i$ and protected attribute $a$:

$$P\{h(x_i)_a = y_i | \ a \ \in A, x_i \in \ X \ \} = P\{h(x'_i)_á = y_i | \ a \ \in A, x_i \in \ X \ \} \qquad (4)$$

This measurement of fairness also provides a mechanism for interpreting the causes of bias because the differences in $h(x_i)$ and $h(x'_i)$ must be caused by variations in $A$. This is because the protected attributes are the only variables that are not controlled.

### 2.5   Embedding Human Reviews in ML Systems

A significant concern when automating decision-making processes using ML systems is that these systems may perpetuate bias against certain groups or subgroups [17, 25–27]. It is believed that incorporating human input and oversight (i.e., human-in-the-loop) in the decision-making process of ML systems can effectively reduce biases present in the system [28]. The reason is that humans bring different perspectives, knowledge, and experiences that the algorithm may not have considered while making automated decisions. Combining the strengths of humans and algorithms can help mitigate potential biases and ensure that the decisions made are fair and unbiased. It also increases accountability and transparency in the decision-making process [29–31].

Human review is a crucial component of human-in-the-loop (HITL) systems, in which human experts are involved in the decision-making process of an ML system. The primary purpose of human review is to provide a final check on the output of the ML model and to ensure that the decisions made by the ML system are fair, unbiased, and accurate. This is usually applied as a retrospective review, also called a post-hoc review. It is an essential step for monitoring and auditing an ML system.

Given $F$ as the function of human review, $ML$ as the function of the ML model, $H$ stands as the judgement by a human expert and $x$ as the input data. The relationship can be expressed as:

$$F\big(ML(x)\big) = \ H\big(ML(x)\big) \ if \ (ML(x)) \ >= \lambda \qquad (5)$$

$F(ML(x))$ represents the system's final decision, either the ML model's decision or the human reviewer's decision. $ML(x)$ represents the ML model's decision for a given input $x$. $H(ML(x))$ represents the human reviewer's decision for a given input $x$. The threshold, $\lambda$, is a pre-defined value that represents the acceptable level of bias.

It can also be defined as a binary function such as:

$$H(ML(x)) = \{1, if\ f(ML(x)) = 1\} \{0, if\ f(ML(x)) = 0\} \qquad (6)$$

For reviewing multiple predictions as in the case of counterfactuals generated, the equation can be rewritten thus:

$$F(ML(x)) = H(ML(x_1), ML(x_2), \ldots, ML(x_n)) \qquad (7)$$

This function could be used to confirm the output of the ML model, by allowing a human to review and verify that the output is correct.

## 3    Related Works

The literature has presented several approaches to identifying individual discrimination in ML systems. Galhotra et al. [32] introduced Themis, which randomly samples the input space to create test cases and assesses the frequency of discriminatory occurrences by observing the system's behaviour under test. Udeshi et al. (Udeshi et al. 2018) presented AEQUITAS, a two-phase search-based fairness testing technique that explores the input space for discriminatory instances and then perturbs the non-protected attributes of instances to generate more discriminatory samples. Aggarwal et al. [33] proposed SG, a search-based test generation method that combines symbolic generation and local explainability to identify discriminatory instances. Fan et al. (Fan et al. 2022) introduced ExpGA, an explanation-guided discriminatory instance generation approach that uses a genetic algorithm to produce many discriminatory instances from seed instances efficiently. Patel et al. (Patel et al. 2022) proposed a method for assessing individual discrimination using combinatorial t-way testing, which constructs an input parameter model from the training dataset and generates test cases to discover fairness violations. These works are based on offline fairness testing, which assesses the fairness of an ML system before its deployment. In contrast, our work focuses on online fairness testing, which evaluates the fairness of an ML system during its operation.

A similar work to our proposed model is BiasRV by Yang et al. [14]. BiasRV is a tool for detecting gender discrimination in sentiment analysis systems. It generates gender-discriminatory mutants from the input text and evaluates the fairness of the system's response using commonly used metrics in sentiment analysis. However, our work differs from theirs in several ways. First, BiasRV is based on text-based ML systems, while our work is based on a tabular dataset. Second, BiasRV uses only gender as the protected attribute, while our work can use multiple attributes as specified by the user. Third, BiasRV uses distributional fairness, while our work uses counterfactual explanations. Lastly, our work includes a human review component to correct any detected discrimination.

## 4 Methodology

In this section, the proposed conceptual model is introduced. The model is designed to monitor and mitigate bias in real time while an ML system is actively functioning, like a loan application system. The process begins with a user inputting data, such as loan application information, which is then fed into the counterfactual generation component, as illustrated in Fig. 1. The components of the model are highlighted as follows.

**Counterfactual generation:** The counterfactual generation process involves inputting an input instance, $x$, sent to the system by a user. To determine whether the instance is discriminatory, the counterfactual explanations engine takes the input instance, a pre-trained ML model, $f$, and a set of protected attributes, $p$, to generate counterfactual examples, $\{c_1, c_2, \dots, c_n\}$, such that they all lead to a different outcome than $x$. The input instance ($x$) and all counterfactual examples have the same features but a different set of protected attributes. The maximum number of counterfactual examples to be generated is based on the different combinations of the values of the protected attributes. This approach provides a more comprehensive evaluation of the model's performance as it considers multiple attributes and not just a single protected attribute. It also helps to uncover subtle biases that may not have been noticeable when analysing the overall population. Furthermore, it enables the assessment of the model's fairness for different subpopulations, which is crucial in ensuring that the model treats all individuals and groups fairly.

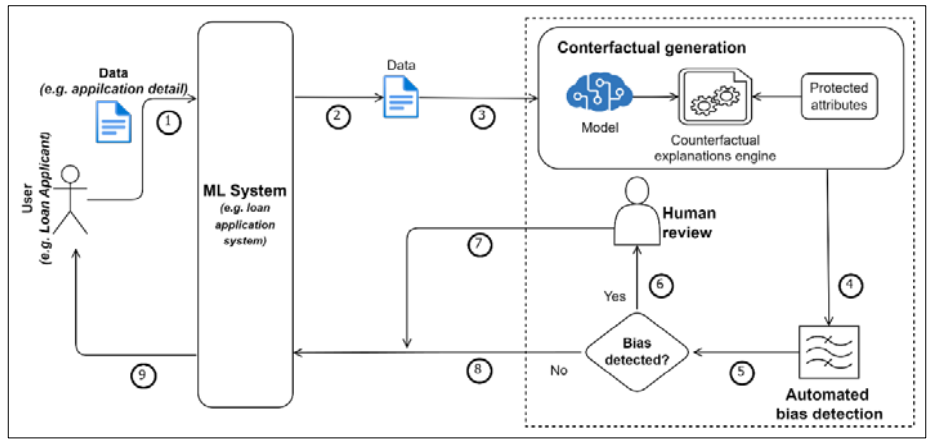

**Fig. 1:** The Proposed Conceptual Model for Real-time Tracking and Correction of Bias

**Bias detection:** The automated bias detection component takes the output from the counterfactual generation process and analyses it to determine whether there is discrimination against the original instance. Suppose the result indicates the presence of counterfactual examples, which have different outcomes from the original sample. In that case, it raises a flag for the human review component to make the final decision. However, if no evidence of discrimination is found (as determined by the absence of counterfactual examples), the ML model's decision stands. The result is recorded and sent to the user.

**Human review:** Human review plays a crucial role in evaluating the fairness of an ML system during operation. The automated bias detection component flags discriminatory instances and generates corresponding counterfactual examples, which are then passed to the human review component for final evaluation. The human review offers an advantage over offline fairness testing by allowing a deeper examination of the reasons behind discriminatory decisions made by the system in real time. The human reviewer can examine multiple subgroups, considering protected attributes, and thereby identify subtle biases, which is crucial in ensuring equitable treatment of all groups. The human reviewer makes a final decision, which may include overturning the model's original decision, which is recorded and communicated to the user. The human review component provides a comprehensive understanding of the model's behaviour and allows for identifying and correcting unfair biases in run-time, leading to a fair and unbiased system.

## 5 Use Case Examples

To demonstrate how our conceptual model can be applied in real-world scenarios, we provide examples in various fields as can be seen in this section.

**Example 1: Healthcare system:** The subject of bias and fairness in healthcare has been a long-standing issue. For example, concerns that clinician biases may perpetuate disparities in healthcare delivery, given the observable and substantial inequalities in treatment approaches for patients of varying race/ethnicity or gender, is well-known [36–38]. Integrating ML systems in healthcare have exacerbated the potential for discriminatory practices and the perpetuation of biased outcomes in some areas [39–42]. Our proposed model can potentially alleviate these disparities in healthcare systems.

Consider a predictive model used to evaluate patients and predict the likelihood of re-admission. The system predicts re-admission risk using patient data such as demographic information, medical history, and lab results. However, if the model makes biased decisions that disproportionately affect certain demographic groups or populations, this can be detected using the conceptual model's counterfactual explanations and automated bias detection components. For instance, the counterfactual explanations could demonstrate that a patient's risk of re-admission would have been lower if they had been a member of a more privileged group.

Another scenario that might take place in healthcare delivery systems involves applying the conceptual model to identify and correct biases that might exist in the patient diagnosis process. For example, one could use a predictive model to evaluate the patient's symptoms and make a recommendation regarding the diagnosis. Suppose the model makes biased decisions disproportionately affecting certain demographic groups or populations. In that case, the conceptual model's counterfactual generation and automated bias detection components can be used to identify biases in the model's decisions. The decisions marked as potentially biased can then be reviewed by a doctor, who can take any necessary steps to correct the bias. In this way, the healthcare system can use the conceptual model to guarantee that all patients receive appropriate and accurate diagnoses regardless of their socioeconomic status.

**Example 2: Education systems:** The proposed conceptual model has applications in the realm of education to monitor and address biases in decision-making processes by admissions committees, scholarship committees, and the like. Certain institutions have been observed to employ race-based factors that result in discriminatory steering of specific demographics towards courses or majors [43]. The model, incorporating counterfactual generation and automated bias detection components, can identify such instances of discrimination by exposing scenarios where, for example, minority group applicants in college admissions may be unfairly directed towards a specific course or major. These biases, once flagged, can be subject to review and correction by the admission committee, thereby promoting a more equitable decision-making process.

**Example 3: Granting load/credit scoring system:** One domain where our conceptual model can be applied is the granting of load. The credit scoring system evaluates applicants for granting or denying a loan. The system may be generating biased decisions that disproportionately reject loan applications from blacks and minority groups [3, 4]. Utilising our framework's counterfactual explanation and automated bias detection components, these biases can be identified and flagged for human review. A human evaluator can then review these flagged decisions to determine the most appropriate credit-scoring decision.

**Example 4: Criminal justice system:** Consider the case of a criminal justice system using an ML system to predict the likelihood of recidivism for defendants. This kind of system has been found to discriminate against black defendants disproportionately [7]. Our proposed model can uncover this discrimination using its counterfactual generation and automated bias detection components. It can show that if a black defendant had been a member of a more privileged (white) group, their risk score would have been lower. When these biases have been identified, justices will be well informed to make appropriate sentencing decisions against defendants.

**Example 5: Hiring system:** Employee selection through an ML system that ranks job applicants based on criteria such as past work experience, education, and abilities has been known to exhibit prejudice against female candidates [1, 2]. Using the proposed model's counterfactual generation and automated bias detection components, these types of biases can be detected during the system's operation. This will reveal instances where a female applicant would have been hired if they were male, providing evidence of biased decision-making by the system, which ultimately limits employment opportunities for female applicants.

These are a few examples where our conceptual model is applicable. The conceptual model has the potential to be used in nearly every domain where automated decision-making relies on tabular data, and there is a risk of discrimination against some subset of the population.

## 6      Conclusions

This paper presents a conceptual model for tracking and correcting individual discrimination in real-time using human review and counterfactual explanations. Our framework leverages the power of counterfactual explanations to pinpoint instances of discrimination in ML systems and includes a human review component to mitigate such biases. This approach helps to guarantee that the decisions made by these systems are fair and unbiased, thereby preventing disadvantaged groups from being unfairly impacted by discriminatory outcomes. Our conceptual model represents a promising approach for addressing biases in deployed ML systems in various domains.

In the future, we plan to bring the proposed system to life by constructing and deploying it. Once the system is up and running, we will follow up with fairness tests to ensure that it operates in a just and equitable manner. These tests will allow us to monitor and address any biases that may arise in real-time, ensuring that the system continues to meet the highest standards of fairness and ethical practice.

## 7      References

1. Dastin, J.: Amazon scraps secret AI recruiting tool that showed bias against women, https://www.reuters.com/article/us-amazon-com-jobs-automation-insight-idUSKCN1MK08G
2. Raghavan, M., Barocas, S., Kleinberg, J., Levy, K.: Mitigating bias in algorithmic hiring: Evaluating claims and practices. In: Proceedings of the 2020 conference on fairness, accountability, and transparency. pp. 469–481 (2020)
3. Hale, K.: A.I. Bias Caused 80% Of Black Mortgage Applicants To Be Denied, https://www.forbes.com/sites/korihale/2021/09/02/ai-bias-caused-80-of-black-mortgage-applicants-to-be-denied/?sh=70038d336feb
4. Counts, C.: Minority homebuyers face widespread statistical lending discrimination, https://phys.org/news/2018-11-minority-homebuyers-widespread-statistical-discrimination.html#google_vignette
5. Mattioli, D.: On Orbitz, Mac users steered to pricier hotels. Wall Street Journal. 23, 2012 (2012)
6. Rajkomar, A., Dean, J., Kohane, I.: Machine learning in medicine. New England Journal of Medicine. 380, 1347–1358 (2019)
7. Angwin, J., Larson, J., Mattu, S., Kirchner, L.: Machine Bias: There's software used across the country to predict future criminals. And it's biased against blacks., https://www.propublica.org/article/machine-bias-risk-assessments-in-criminal-sentencing
8. Joh, E.E.: Feeding the machine: Policing, crime data, & algorithms. Wm. & Mary Bill Rts. J. 26, 287 (2017)
9. Pessach, D., Shmueli, E.: A Review on Fairness in Machine Learning. ACM Comput Surv. 55, 1–44 (2022). https://doi.org/10.1145/3494672
10. Quy, T. Le, Roy, A., Iosifidis, V., Zhang, W., Ntoutsi, E.: A survey on datasets for fairness-aware machine learning. 1–56 (2021)

11. Brun, Y., Meliou, A.: Software fairness. ESEC/FSE 2018 - Proceedings of the 2018 26th ACM Joint Meeting on European Software Engineering Conference and Symposium on the Foundations of Software Engineering. 754–759 (2018). https://doi.org/10.1145/3236024.3264838
12. Udeshi, S., Arora, P., Chattopadhyay, S.: Automated directed fairness testing. ASE 2018 - Proceedings of the 33rd ACM/IEEE International Conference on Automated Software Engineering. 98–108 (2018). https://doi.org/10.1145/3238147.3238165
13. Monjezi, V., Trivedi, A., Tan, G., Tizpaz-Niari, S.: Information-Theoretic Testing and Debugging of Fairness Defects in Deep Neural Networks. arXiv preprint arXiv:2304.04199. (2023)
14. Yang, Z., Asyrofi, M.H., Lo, D.: BiasRV: Uncovering biased sentiment predictions at runtime. ESEC/FSE 2021 - Proceedings of the 29th ACM Joint Meeting European Software Engineering Conference and Symposium on the Foundations of Software Engineering. 1, 1540–1544 (2021). https://doi.org/10.1145/3468264.3473117
15. Chen, Z., Zhang, J.M., Hort, M., Sarro, F., Harman, M.: Fairness Testing: A Comprehensive Survey and Analysis of Trends. (2022)
16. Maughan, K., Ngong, I.C., Near, J.P.: Prediction Sensitivity: Continual Audit of Counterfactual Fairness in Deployed Classifiers. In: arXiv preprint arXiv:2202.04504. IEEE Computer Society (2022)
17. Mehrabi, N., Morstatter, F., Saxena, N., Lerman, K., Galstyan, A.: A Survey on Bias and Fairness in Machine Learning. ACM Comput Surv. 54, (2021). https://doi.org/10.1145/3457607
18. Dwork, C., Hardt, M., Pitassi, T., Reingold, O., Zemel, R.: Fairness through awareness. ITCS 2012 - Innovations in Theoretical Computer Science Conference. 214–226 (2012). https://doi.org/10.1145/2090236.2090255
19. Tramèr, F., Atlidakis, V., Geambasu, R., Hsu, D., Hubaux, J.P., Humbert, M., Juels, A., Lin, H.: FairTest: Discovering Unwarranted Associations in Data-Driven Applications. Proceedings - 2nd IEEE European Symposium on Security and Privacy, EuroS and P 2017. 401–416 (2017). https://doi.org/10.1109/EuroSP.2017.29
20. Zhang, J.M., Harman, M., Ma, L., Liu, Y.: Machine Learning Testing: Survey, Landscapes and Horizons. IEEE Transactions on Software Engineering. 48, 1–36 (2022). https://doi.org/10.1109/TSE.2019.2962027
21. Wachter, S., Mittelstadt, B., Russell, C.: Counterfactual explanations without opening the black box: Automated decisions and the GDPR. Harv. JL & Tech. 31, 841 (2017)
22. Mothilal, R.K., Sharma, A., Tan, C.: Explaining machine learning classifiers through diverse counterfactual explanations. In: FAT* 2020 - Proceedings of the 2020 Conference on Fairness, Accountability, and Transparency. pp. 607–617. Association for Computing Machinery, Inc (2020)
23. Kusner, M.J., Loftus, J., Russell, C., Silva, R.: Counterfactual fairness. Adv Neural Inf Process Syst. 30, (2017)
24. Black, E., Yeom, S., Fredrikson, M.: FlipTest: Fairness testing via optimal transport. FAT* 2020 - Proceedings of the 2020 Conference on Fairness,

Accountability, and Transparency. 111–121 (2020). https://doi.org/10.1145/3351095.3372845

25. Balayn, A., Lofi, C., Houben, G.J.: Managing bias and unfairness in data for decision support: a survey of machine learning and data engineering approaches to identify and mitigate bias and unfairness within data management and analytics systems. VLDB Journal. 30, 739–768 (2021). https://doi.org/10.1007/s00778-021-00671-8
26. Chakraborty, J., Peng, K., Menzies, T.: Making Fair ML Software using Trustworthy Explanation. Proceedings - 2020 35th IEEE/ACM International Conference on Automated Software Engineering, ASE 2020. 1229–1233 (2020). https://doi.org/10.1145/3324884.3418932
27. Li, B., Qi, P., Liu, B., Di, S., Liu, J., Pei, J., Yi, J., Zhou, B.: Trustworthy AI: From Principles to Practices. 1, (2021)
28. Cheng, H.-F.: Advancing Explainability and Fairness in AI with Human-Algorithm Collaborations, (2022)
29. Elmalaki, S.: FaiR-IoT: Fairness-aware Human-in-the-Loop Reinforcement Learning for Harnessing Human Variability in Personalized IoT. In: IoTDI 2021 - Proceedings of the 2021 International Conference on Internet-of-Things Design and Implementation. pp. 119–132. Association for Computing Machinery, Inc (2021)
30. Mosqueira-Rey, E., Hernández-Pereira, E., Alonso-Ríos, D., Bobes-Bascarán, J., Fernández-Leal, Á.: Human-in-the-loop machine learning: a state of the art. Artif Intell Rev. (2022). https://doi.org/10.1007/s10462-022-10246-w
31. Wu, X., Xiao, L., Sun, Y., Zhang, J., Ma, T., He, L.: A survey of human-in-the-loop for machine learning, (2022)
32. Galhotra, S., Brun, Y., Meliou, A.: Fairness testing: Testing software for discrimination. Proceedings of the ACM SIGSOFT Symposium on the Foundations of Software Engineering. Part F1301, 498–510 (2017). https://doi.org/10.1145/3106237.3106277
33. Aggarwal, A., Lohia, P., Nagar, S., Dey, K., Saha, D.: Black box fairness testing of machine learning models. ESEC/FSE 2019 - Proceedings of the 2019 27th ACM Joint Meeting European Software Engineering Conference and Symposium on the Foundations of Software Engineering. 625–635 (2019). https://doi.org/10.1145/3338906.3338937
34. Fan, M., Wei, W., Jin, W., Yang, Z., Liu, T.: Explanation-Guided Fairness Testing through Genetic Algorithm. In: 2022 IEEE/ACM 44th International Conference on Software Engineering (ICSE). pp. 871–882 (2022)
35. Patel, A.R., Chandrasekaran, J., Lei, Y., Kacker, R.N., Kuhn, D.R.: A Combinatorial Approach to Fairness Testing of Machine Learning Models. (2022)
36. Williams, D.R., Wyatt, R.: Racial Bias in Health Care and Health (Reprinted) JAMA August 11. (2015)
37. Wong, W.F., LaVeist, T.A., Sharfstein, J.M.: Achieving health equity by design, (2015)
38. Wailoo, K.: Historical Aspects of Race and Medicine: The Case of J. Marion Sims, (2018)

39. Ahmad, M.A., Patel, A., Eckert, C., Kumar, V., Teredesai, A.: Fairness in Machine Learning for Healthcare. In: Proceedings of the ACM SIGKDD International Conference on Knowledge Discovery and Data Mining. pp. 3529–3530. Association for Computing Machinery (2020)
40. Paulus, J.K., Kent, D.M.: Predictably unequal: understanding and addressing concerns that algorithmic clinical prediction may increase health disparities. NPJ Digit Med. 3, 1–8 (2020). https://doi.org/10.1038/s41746-020-0304-9
41. Norori, N., Hu, Q., Aellen, F.M., Faraci, F.D., Tzovara, A.: Addressing bias in big data and AI for health care: A call for open science. Patterns. 2, 100347 (2021). https://doi.org/10.1016/j.patter.2021.100347
42. Grote, T., Keeling, G.: Enabling Fairness in Healthcare Through Machine Learning. Ethics Inf Technol. 24, (2022). https://doi.org/10.1007/s10676-022-09658-7
43. Feathers, T.: Major Universities Are Using Race as a "High Impact Predictor" of Student Success. In: Ethics of Data and Analytics. pp. 268–273. Auerbach Publications (2022)